\documentclass[
]{ceurart}

\usepackage{listings}
\usepackage[svgnames]{xcolor}
\begin{document}

\copyrightyear{2024}
\copyrightclause{Copyright for this paper by its authors.
  Use permitted under Creative Commons License Attribution 4.0
  International (CC BY 4.0).}

\conference{GeoExT 2024: Second International Workshop on Geographic Information Extraction from Texts at ECIR 2024, March 24, 2024, Glasgow, Scotland}

\title{Exploring Spatial Representations in the Historical Lake District Texts with LLM-based Relation Extraction}

\tnotemark[]

\author[1]{Erum Haris}[%
orcid=0000-0002-8012-8850,
email=e.haris@leeds.ac.uk,
]
\cormark[1]
\address[1]{School of Computing, University of Leeds, UK}

\author[1,2]{Anthony G. Cohn}[%
orcid=0000-0002-7652-8907,
email=a.g.cohn@leeds.ac.uk,
] 
\address[2]{The Alan Turing Institute, London, UK}

\author[1]{John G. Stell}[%
orcid=0000-0001-9644-1908,
email=j.g.stell@leeds.ac.uk,
]

\cortext[1]{Corresponding author.}

\begin{abstract}
  Navigating historical narratives poses a challenge in unveiling the spatial intricacies of past landscapes. The proposed work addresses this challenge within the context of the English Lake District, employing the Corpus of the Lake District Writing. The method utilizes a generative pre-trained transformer model to extract spatial relations from the textual descriptions in the corpus. The study applies this large language model to understand the spatial dimensions inherent in historical narratives comprehensively. The outcomes are presented as semantic triples, capturing the nuanced connections between entities and locations, and visualized as a network, offering a graphical representation of the spatial narrative. The study contributes to a deeper comprehension of the English Lake District's spatial tapestry and provides an approach to uncovering spatial relations within diverse historical contexts.
\end{abstract}


\maketitle
\section{Introduction}

Recent years have witnessed a growing interest in understanding geographies within textual sources across various disciplines. In the realm of historical archives, the Corpus of the Lake District Writings (CLDW) \cite{Donaldson17} has attracted many researchers to explore the English Lake District, a region known for its cultural heritage, characterized by lakes and mountains in the northwest of England. The collection contains spatial narratives with descriptions of landscapes, places and routes. The writers' experiences of geography, as depicted in these narratives, are subjective and a detailed analysis is necessary to handle the qualitative aspects of spatial information. 

A particular case is the representation of relative locations in narratives using qualitative spatial relations. References such as ``near'', ``to the south'', or ``a few hours drive from'' are often used by writers as they experience relative locations. However, owing to the ambiguity of natural language, it is a challenge to precisely extract spatial relations between any two entities from text data. Spatial relations are classified into different types \cite{Haris20}, including topological, direction and distance relations ; each category requires a contextual understanding of the text for extraction. Moreover, relation arity, toponym matching, coreference resolution, long-dependency relations and related issues make relation extraction, in general, a demanding research problem. 

Spatial relation extraction has been primarily addressed either using manually crafted rules and patterns to be matched over sentence tokens or using supervised learning, which requires an enormous amount of annotated training data. Web-scale open information extraction \cite{Detroja23}, though, does not require patterns in advance, but it results in many redundant and incoherent facts in output. The significant breakthroughs driven by Large Language Models (LLMs) \cite{Brown20} in processing natural language text have led to potential use cases and research findings in GIS and spatial analysis. LLMs are pre-trained on large-scale textual data, which enables them to generalize to unseen data without annotated examples or requiring minimal examples. Numerous studies have concluded their success in various downstream NLP tasks \cite{Wei23,Yuan23}; hence, they are also called foundation models \cite{Hu23}.

In this paper, we integrate two problem domains and perform a holistic analysis. We aim to unravel the spatial representations in the CLDW corpus using a Generative Pre-trained Transformer (GPT) model \cite{Brown20}, thereby exploring the potential of pre-trained models in extracting spatial relations from historical accounts. Specifically, the experiment focuses on the extraction of the spatial relation ``near'' between entities, and results are visualized as a network of semantic triples.

\section{Related Work}

The proposed study overlaps with multiple areas, including GIS and NLP in spatial humanities, particularly studies on the CLDW, spatial relation extraction and applications of LLMs in geographical information science (GISc). Hence, a brief review is presented for each domain. 

Donaldson et al. \cite{Donaldson17} propose an interdisciplinary methodology for analyzing historical text corpora. The study integrates corpus analysis, automated geoparsing and GIS to examine the spatial dimensions linked to significant aesthetic terms historically used to depict the English Lake District. The approach, termed ``geographical text analysis,'' demonstrates its capacity to advance the analysis of interconnections among literature, aesthetics, and physical geography. Going beyond toponymic geography, Ezeani et al. \cite{Ezeani23} introduce an extensible framework to study the CLDW using NLP, GISc, qualitative spatio-temporal representation and reasoning (QSTR), and visual analytics. They demonstrate preliminary work using the framework on extracting, analyzing, and visualizing the spatial elements that define the ``location'', ``locale'' and ``sense of place'' as referenced in the text.  

Proximity holds a pivotal role in any comprehensive spatial ontology, \cite{Worboys01,Brennan09}.  The spatial concept of nearness or proximity serves to identify either the spatial relationship between two entities, such as ``near (street, village),'' or an object's relative location in space ``street (near, village),'' encapsulating people's psychological perceptions of distances \cite{Novel20}. However, interpreting the meaning of qualitative proximity statements remains challenging within GIS. Talmy \cite{Talmy83} describes that natural language expressions representing spatial relations often exhibit asymmetry. Along this line, Worboys \cite{Worboys01} explores nearness relations in environmental spaces, considering nearness as a ``conceptual'' distance relation. Novel et al. \cite{Novel20} propose a context-dependent model for defining ``near,'' leveraging Turney's two dimensional contextualization framework \cite{Turney02}.

In spatial relation extraction, methods usually focus on a single aspect of spatial information, or sometimes a few aspects, owing to the domain-dependent nature of the corpus language. Drymonas and Pfoser \cite{Drymonas10} reconstruct route maps by extracting relative and absolute geospatial information from travel guides using a rule-based extraction module and a location ontology. Skoumas et al. \cite{Skoumas15} extract short route indicators including distance and topological relations from crowdsourced geospatial data using custom-defined regular expression patterns and syntactic rules. Cadorel et al. \cite{Cadorel21} build a structured Geospatial knowledge base with a spatial information extraction pipeline including named entities and relations developed using recurrent neural networks.  

The application of LLMs in the geospatial domain has resulted in various research studies. Hu et al. \cite{Hu23} integrate geo-knowledge from location descriptions with a GPT model resulting in a geo-knowledge-guided GPT model, designed explicitly for extracting precise location information from social media messages related to disasters. Mooney et al. \cite{Mooney23} scrutinize the capabilities of ChatGPT in specialized subject area like GIS, acknowledging the significant divergence between human learning of spatial concepts and LLM training methodologies. The evaluation entails subjecting ChatGPT to a real GIS exam assessing its performance and spatial literacy. Addressing one of the facets of spatial relations, Ramrakhiyani et al. \cite{Ramrakhiyani23} focus on extracting the orientation information of borders between countries from Wikipedia text using large language models based on Natural Language Inference (NLI) technique, combined with lexical patterns.  

\section{Data and Method}
\subsection{Corpus of the Lake District Writing}

This work is part of a spatial exploration of the CLDW \cite{Rayson17}, a collection encompassing travel writing and tourist literature that vividly depicts the English Lake District from 1622 to 1900. Comprising 80 geoparsed texts with over 1.5 million words and encompassing various genres, including travel writing and fiction, the corpus provides valuable accounts from the Lake Poets like Wordsworth and Coleridge and other significant figures, alongside works from lesser-known authors. The descriptions capture the essence of travelling, aiming to describe landscapes and emotional responses. 

We considered the instances of the term ``near'' in the corpus for this study. Table~\ref{tab:stats} presents a statistical summary of the spatial content in the CLDW, where named place refers to any occurrence of a toponym and geographic noun represents a geographical feature appearing in the text. A list of 139 nouns (e.g. river, road, waterfall, etc) as well as their inflections (rivers, roads, waterfalls) has been manually curated[8]. Spatial references are the different occurrences of spatial relations, including spatial prepositions, locative adverbs and distances \cite{Ezeani23}. Table~\ref{tab:contx} lists keyword-in-context (KWIC) \cite{Voyant16} examples for the relation ``near'' showing its left and right context within text passages. 

\begin{table}
  \caption{Statistical information on spatial content of the CLDW}
  \label{tab:stats}
  \begin{tabular}{lr}
    \toprule
    Parameter & Count \\
    \midrule
    \texttt Text Files & 80 \\
    \texttt Words & 1,524,718 \\
    \texttt Unique Word Forms & 43,765 \\
    \texttt Named Places & 39,916 \\
    \texttt Geographic Nouns & 69,103 \\
    \texttt Spatial References & 41,901 \\
    \texttt "Near" Relation Occurrences  & 1611 \\
    \bottomrule
  \end{tabular}
\end{table}

\begin{figure}
  \centering
  \includegraphics[width=\linewidth]{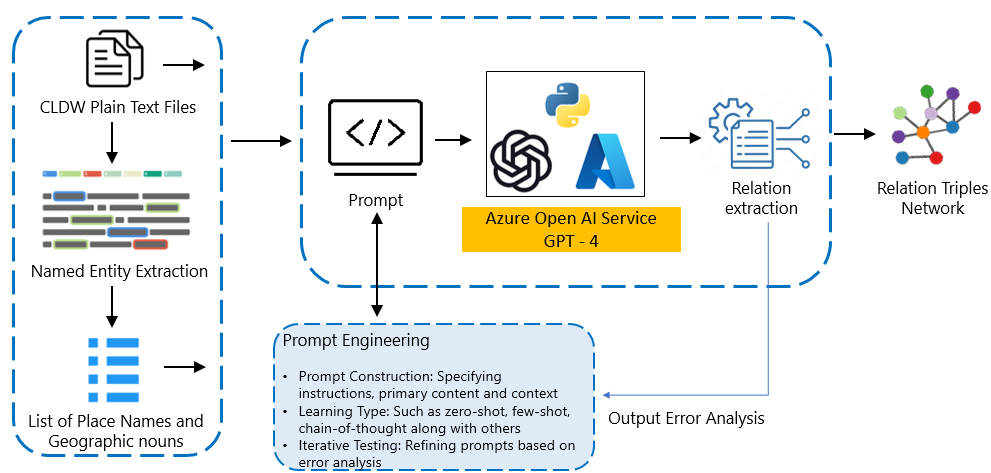}
  \caption{Proposed framework for relation extraction and visualization. Template style (\url{https://learn.microsoft.com/en-us/azure/architecture/ai-ml/guide/query-based-summarization}).}
\end{figure}

\subsection{Framework}
Figure 1 illustrates the overall framework for relation extraction, which consists of two stages. The first is the entity extraction stage, which provides place names and geographic nouns in the CLDW using the named entity recognition (NER) module developed by Ezeani et al. \cite{Ezeani23}. These lists and the plain corpus files are used to design prompts for the pre-trained model, details of which are described in the experiment section. The method employs GPT-4 using the Azure OpenAI services \cite{Azure23}, which provide access to OpenAI's powerful language models, some of which have reached general availability and can be deployed to NLP and image understanding. The prompts are developed and passed to the model through Python SDK. The resultant semantic triples are compared with the gold-standard set of labelled sentences for precision calculation. Finally, the correct triples are passed to the network visualization module for a graphical illustration of the target place's spatial connections.

\subsection{LLM Prompt Engineering for Relation Extraction}

LLMs are powerful but can sometimes generate outputs not aligned with the specific intent. Hence, prompt engineering takes effect \cite{Liu23}, which refers to the deliberate design and formulation of prompts or queries provided to LLMs, particularly in fine-tuning or guiding the behaviour of these models to improve their usability and reliability. In practical terms, the prompt serves as a mechanism that directs the model's attention and influences its output. The model adjusts its internal weights and parameters during the training or inference process \cite{Azure23}. This adjustment aligns the model's behavior to better suit the intended task, improving its performance and accuracy in generating responses or predictions. A model’s behaviour is highly sensitive to the prompt; therefore, formulating prompts is challenging and requires considerable effort.

\begin{table}
  \caption{Keyword-in-context examples for spatial relation ``near''}
  \label{tab:contx}
  \begin{tabular}{rcl}
    \toprule
    Left & Term & Right \\
    \midrule
    \texttt station is at Coniston village & near & the head of the lake\\
    \texttt seen in the church-yard & near & the road. We have now \\
    \texttt entered among the luxuriant woods & near & Gowbarrow Park and dark Ulleswater's\\
    \texttt in the coves of Wythburn & near & to Dunmail Raise, and, after\\
    \bottomrule
  \end{tabular}
\end{table}

In our approach to spatial relation extraction, the prompt engineering stage essentially follows these steps. The process starts with specifying the task instructions and giving content and context to assist the model's understanding, defining the desired output format and experimenting with different prompt structures without providing training examples. This method is called zero-shot learning \cite{Yuan23}, in which the model works to correlate the provided task with existing categories it has previously learned and formulates responses accordingly. Finally, prompt engineering is an iterative process that involves testing and refining prompts based on the model's output errors until the desired results are achieved. We performed several iterations following a top-down approach, starting with open-ended spatial relation extraction to confining to only specific category of spatial relations, extracting relations as predicates and then spatial prepositions, mentioning spatial entities to be looked for, providing single and multiple paragraphs and as a minimum, changing instruction articulation in the prompt. After performing various experiments, we finally resorted to starting with the most basic setting for extracting spatial relations to see which places are nearby located to others. Hence, the term ``near'' is chosen at first for extraction and analysis. It is apparent that terms like ``close to'', ``surrounds'', ``next to'' and others also convey a sense of proximity. However, in this work, we have explicitly focused on the keyword ``near'' to formulate the prompts at this stage better. We describe the technique followed in the experiment section.

\section{Experiment and Results}

Efficient crafting of input prompts must consider the token count constraint, which influences the model's response latency and throughput \cite{Azure23}.  Therefore, the corpus was first subjected to some text-mining techniques before passing the text to the model. First, a KWIC analysis was performed for the spatial relation ``near'' with a context window of 15 words before and after, which is the maximum allowed window size in the text analytics software \cite{Max22}. The search resulted in 1609 contexts. As the research is in the initial phases of assessing the GPT model's performance, we attempted to refine the search further. Our next choice was to look for the most recurring places in the list obtained from the NER stage. Hence, a frequency count of the place names was retrieved, according to which the town of ``Keswick'' is the most frequently occurring place name. Applying the filtering with the keyword ``Keswick'' on the relation ``near'' contexts' list resulted in 84 outcomes. Table~\ref{tab:precision} provides information on this analysis for the most frequent place names in the CLDW.

\subsection{Zero-shot Spatial Relation Extraction}
Table~\ref{tab:prompt} gives a glimpse of the descriptive prompt provided to the GPT-4 model and the generated response. We explicitly specified three things inside the prompt in this zero-shot setting:

\begin{itemize}
\item \verb|()| : the numbered appearance of the input passages
\item \verb|near|: the spatial relation to look for
\item \verb|<subject, spatial relation, object>|: the format of the output where ``subject'' and ``object'' are the spatial entities, one of which must be the specified place name which is ``Keswick'' in this context
\end{itemize}

\begin{table}
  \caption{Zero-shot Prompt with response}
  \label{tab:prompt}
  \begin{tabular}{p{3.5in}|p{1in}}\hline
    \toprule
    Prompt & Response\\
    \midrule
    \texttt {\color{LimeGreen} "content":"From the given passages numbered in brackets (), extract spatial relation "near" only if the entity 'keswick' is involved in a relation with other entities. The output should be in the form <subject, spatial relation, object>.(1)Carlisle railway conducts from Kendal to Birthwaite, a mile to the north of Bowness, and near the shores of the lake. From the station here coaches run to Ambleside and Keswick (2) is not to be found in Cumberland or Westmoreland. The view is taken from Castlehead, near Keswick. The Index of Places will enable the reader to turn immediately to the page (3) on the small scale is the loveliest landscape in nature. Southey lives at Greta Hall, near Keswick, with Skiddaw at his back, and Derwentwater spread out before him. Coleridge, who married... "} & {\color{blue} "content": "(1) <Castlehead, near, Keswick>(2) <Greta Hall, near, Keswick>(3) <Scale Hill, near, Keswick>(4) <Naddle, near, Keswick>(5) <Castrigg, near, Keswick>(6) <Lowdore, near, Keswick>(7) <Grange Bridge, near, Keswick>...}\\
    \bottomrule
  \end{tabular}
\end{table}

We are interested in using the generalization power of the GPT model and assessing its overall effectiveness in identifying the spatial relationships between entities in the text without providing specific examples. This is desirable in cases where input data is very large and pre-annotation incurs a high cost, as is the case of the CLDW. Hence, in the initial stages, it is preferred that the model should achieve good precision on its task, which means that extracted facts should be correct in comparison with total predictions, though not complete, as we have not guided the model with example use-cases which can cover different patterns of target spatial relation occurrences. On the other hand, the model may not provide high-precision results since in zero-shot learning, the model leverages statistical patterns learned during its training to determine what should be generated next and hence, it may not have a semantic understanding of the problem \cite{Azure23}. 

Table~\ref{tab:precision} provides precision scores for extracting ``near'' spatial relation in different cases. The resultant triples were compared with a gold standard set of sentences labeled with relation occurrences and corresponding place names. Hence, the precision scores denote the fraction of correct triples out of all the triples produced. The prompt was repeated for two iterations to ensure the model generates consistent results, and average precision was computed. The results show that the model can perform in a zero-shot setting with better precision. However, the underlying textual details are highly influential. In many cases, the sentence structure was not straight enough for the model to precisely extract relation predicates. The model was also unable to interpret metaphorical references in some instances correctly. Moreover, the geographical descriptions vary from one type of entity to another, which was the case of ``Windermere'', a lake, and its features and nearby landmarks have been described in comparatively complex phrases. Similarly, ``Ambleside'' is primarily expressed in the context of roads around it. Despite the textual nuances, the model tried to infer nearness in some cases, representing its ability beyond explicit extraction.

It is imperative to mention that we specified the output format to the model in a zero-shot setting. Without specification, prompting for extracting ``near'' relation between entities resulted in large chunks of words extracted as subject or object, merging relation term with object and other unexpected output forms. Next, there is a possibility that the terms ``near'' and ``Keswick'', in this scenario, do not hold any relation despite being present in the filtered context. In such a case, the model may try to fabricate output that can be matched and pointed out but would reduce the overall precision if left untreated. Hence, we also tried adding an instruction in the prompt to respond ``not found'' if the model is not confident about the predicted relation. This attempt resulted in a downgraded response because the model discarded most of the correct triples it had previously extracted. Finally, the environment-specific parameters play crucial role in generating deterministic responses. In Azure OpenAI Service, the prompt is passed to a completion endpoint with other probability parameters, such as ``temperature'' which has been kept to zero to achieve replicable results \cite{Azure23}.

\begin{figure}
  \centering
  \includegraphics[width=\linewidth]{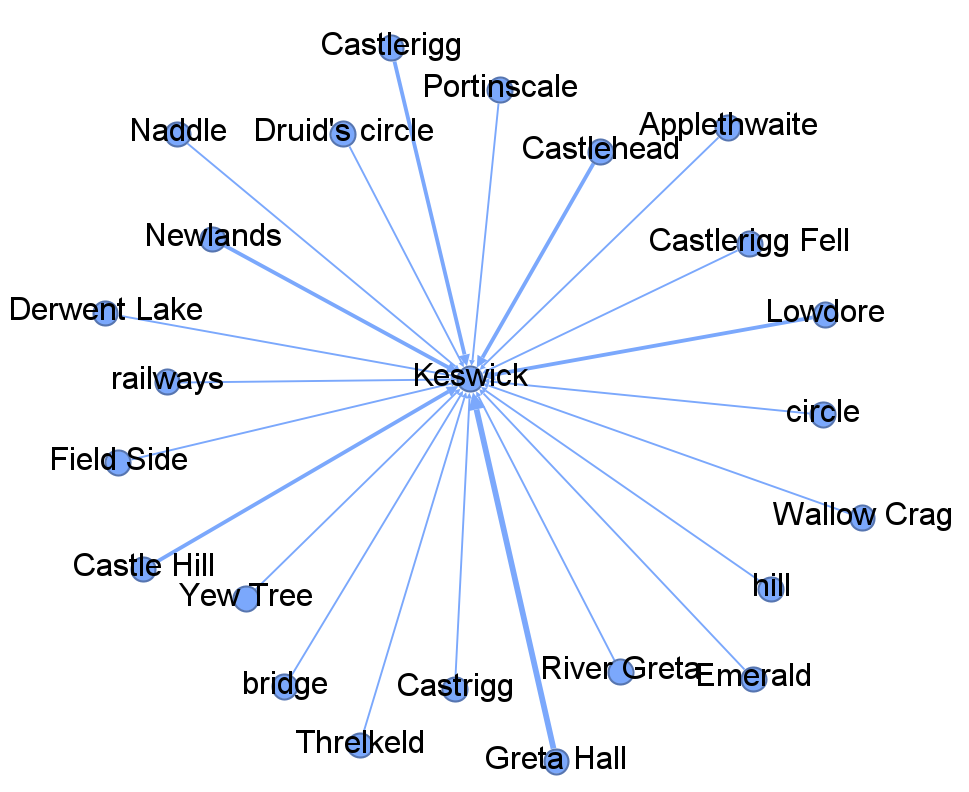}
  \caption{Network visualization of ``near''semantic triples for place``Keswick''.}
\end{figure}

\subsection{Network of Semantic Triples}
For visualization, incorrect outputs were filtered from the model's response. The remaining were used to construct a network \cite{Bastian09}, shown in Figure 2, that visualizes the resultant semantic triples for the place name ``Keswick''. The nodes represent all places in resultant triples without toponym matching to merge nodes referring to the same place. The edges are weighted according to their output frequency in the triples set. Some relation instances occur more than once than others reflected via thickened edges. 

As described earlier, ``near'' is a vague term used subjectively. In most cases, the nearness implies genuine proximity. Still, some destinations requiring more than two or three hours of walking may be related as near to the source depending on the mode of travel, which should be identified from textual descriptions. To strike a balance in our analysis, we explicitly checked the proximity of place names when there was an ambiguity while labeling their relation. This was needed owing to the expressions of nearness in the text challenging the gold standard preparation.

\begin{table}
  \caption{Precision results for four selected places}
  \label{tab:precision}
  \begin{tabular}{lccl}
    \toprule
    Place name & Frequency & Context Count with ``Near''& Average 
 Precision\\
    \midrule
    \texttt Keswick & 1452 & 84 & (0.630, 0.702) = 0.666\\
    \texttt Ambleside & 900 & 43 & (0.656, 0.647) = 0.652\\
    \texttt Windermere & 873 & 45 & (0.543, 0.675) = 0.609\\
    \texttt Penrith & 714 & 53 & (0.714, 0.606) = 0.660\\
    \bottomrule
  \end{tabular}
\end{table}

\section{Conclusion}
 This paper is part of an overarching research initiative that seeks to unveil the spatial and temporal dynamics inherent in narratives \cite{SpaceTime23}. The study aims to introduce a considerable shift in our methodical approaches to exploring the geographies outlined in extensive historical archives by harnessing the power of LLMs. We proposed a framework for extracting spatial relations from the CLDW and presented results for relation ``near''. The results are visualized as a network that depicts the target place, showing its nearby spatial entities. The proposed approach complements the existing geographical analyses by introducing a distinctive computational representation of place, thereby enhancing the capacity of social scientists and humanists to interpret narrative depictions of location. For the future, we are working towards improving the extraction performance by refining the zero-shot prompts, experimenting with few-shot learning and extracting other qualitative spatial relations. 

\begin{acknowledgments} 
  We acknowledge the support of the Economic and Social Research Council (ESRC) under grant ES/W003473/1. The support of Microsoft under their Accelerating Foundation Models Research initiative in providing Azure credits to AGC is also acknowledged. Finally, we thank Paul Rayson and Ignatius Ezeani for their comments on this work, and the entire team of the Spatial Narratives project for their discussions on the CLDW. 
\end{acknowledgments}

\bibliography{sample-ceur}

\appendix

\end{document}